\documentclass[letterpaper, 10 pt, conference]{ieeeconf}

\usepackage{textcomp}
\usepackage{soul}
\usepackage[hidelinks]{hyperref}
\usepackage{caption}

\IEEEoverridecommandlockouts  

\usepackage{fix-cm}
\usepackage{etex}

\usepackage{dblfloatfix}

\usepackage{nag}

\makeatletter
\@ifpackageloaded{xcolor}{}{%
\usepackage[table,x11names,dvipsnames,svgnames]{xcolor}%
}
\makeatother

\usepackage{colortbl}

\usepackage{graphicx}
\usepackage{wrapfig}

\definecolorset{RGB}{lyft}{}{Red,194,39,36;Sunset,202,53,33;Orange,205,68,20;Amber,200,117,42;Yellow,242,169,52;Citron,186,188,44;Lime,112,159,33;Green,56,139,31;Mint,45,118,56;Teal,52,133,135;Cyan,60,132,202;Blue,55,94,248;Indigo,64,13,247;Purple,115,42,248;Pink,176,25,145;Rose,176,32,75}

\usepackage{cite}

\usepackage{microtype}

\usepackage{array}
\usepackage{multirow}
\usepackage{booktabs}
\usepackage{makecell} 

\newcommand{\myparagraph}[1]{\textbf{\emph{#1}}.}

\ifcsname labelindent\endcsname
\let\labelindent\relax
\fi
\usepackage[inline]{enumitem}

\usepackage{subfig}

\newenvironment{lenumerate}[2][]
{\begin{enumerate}[label=(#2\arabic*),leftmargin=0.2in,itemindent=0.15in,#1]}
{\end{enumerate}}

\setlist*[enumerate,1]{label={\itshape\arabic*)}}

\makeatletter
\newcommand{\paragraphswithstop}{%
\let\copyparagraph\paragraph%
\renewcommand\paragraph[1]{\copyparagraph{##1.}}%
}
\makeatother

\usepackage[framemethod=tikz]{mdframed}

\usepackage{suffix}

\usepackage{environ}

\makeatletter
\newsavebox{\boxifnotempty}
\newcommand{\displayifnotempty}[3]{\sbox\boxifnotempty{#2}\setbox0=\hbox{\usebox{\boxifnotempty}\unskip}%
\ifdim\wd0=0pt
\else
 #1\usebox{\boxifnotempty}#3%
\fi%
}
\newcommand{\ifempty}[2]{\setbox0=\hbox{#1\unskip}%
\ifdim\wd0=0pt%
 #2%
\fi%
}
\newcommand{\ifnotempty}[2]{\setbox0=\hbox{#1\unskip}%
\ifdim\wd0>0pt%
 #2%
\fi%
}
\makeatother

\usepackage{algpseudocode}
\usepackage{algorithm}

\usepackage{scrlfile}

\makeatletter
\newcommand*\newstoreddef[1]{
  \BeforeClosingMainAux{%
    \immediate\write\@auxout{%
      \string\restoredef{#1}{\csname #1\endcsname}%
    }%
  }%
}
\newcommand*{\restoredef}[2]{
  \expandafter\gdef\csname stored@#1\endcsname{#2}%
}
\newcommand*{\storeddef}[1]{
  \@ifundefined{stored@#1}{0}{\csname stored@#1\endcsname}%
}
\makeatother

\usepackage{pageslts}
\NewEnviron{tee}{\BODY\typeout{Marker Tee [start] ^^J \BODY ^^JMaker Tee [end]}}

\input{preamble/math}
\newcommand{\real}[1]{\mathbb{R}^{#1}{}}

\providecommand{\cC}{\mathcal{C}}

\providecommand{\cF}{\mathcal{F}}

\providecommand{\cK}{\mathcal{K}}
\providecommand{\cL}{\mathcal{L}}
\providecommand{\cM}{\mathcal{M}}
\providecommand{\cN}{\mathcal{N}}

\providecommand{\cR}{\mathcal{R}}
\providecommand{\cS}{\mathcal{S}}

\usepackage{units}

\newcommand{\newcolorlabel}[2]{%
  \expandafter\newcommand\csname #1\endcsname[1]{%
    \colorbox{#2}{\color{white}\textsf{\textbf{##1}}}}%
}

\newcommand{\newcommenter}[2]{%
  \expandafter\newcommand\csname #1\endcsname[1]{%
    \fcolorbox{#2}{#2}{\color{white}\textsf{\textbf{#1}}}
    {\color{#2}##1}}%
  \expandafter\newcommand\csname at#1\endcsname{%
    \fcolorbox{#2}{#2}{\color{white}\textsf{\textbf{@#1}}}
    {\color{#2}}}%
  \expandafter\newcommand\csname #1hl\endcsname[2]{%
    \colorbox{#2}{\color{white}\textsf{\textbf{#1}}}\sethlcolor{Azure2}\hl{##2}~%
    \expandafter\ifx\csname commentarrow\endcsname\relax$\leftarrow$\else \commentarrow[#2]\fi~%
    {\color{#2}##1}}%
  \expandafter\newcommand\csname #1st\endcsname[2]{%
    \colorbox{#2}{\color{white}\textsf{\textbf{#1}}}\sout{##2}~%
    \expandafter\ifx\csname commentarrow\endcsname\relax$\leftarrow$\else \commentarrow[#2]\fi~%
    {\color{#2}##1}}%
}
\newcommenter{TODO}{DodgerBlue1}
\newcommenter{rtron}{Green3}

\usepackage{comment}

\usepackage{pdfcomment}

\usepackage{soul}

\usepackage[normalem]{ulem}

\usepackage{csquotes}

\usepackage{tikz}
\usetikzlibrary{calc}
\usetikzlibrary{matrix}
\usetikzlibrary{chains}
\usetikzlibrary{shapes.geometric}
\usetikzlibrary{arrows.meta}
\usetikzlibrary{decorations.pathreplacing}
\usetikzlibrary{backgrounds}

\tikzset{
  dim above/.style={to path={\pgfextra{
        \pgfinterruptpath
        \draw[>=latex,|->|] let
        \p1=($(\tikztostart)!1.5em!90:(\tikztotarget)$),
        \p2=($(\tikztotarget)!1.5em!-90:(\tikztostart)$)
        in(\p1) -- (\p2) node[pos=.5,sloped,above]{#1};
        \endpgfinterruptpath
      }
    }
  },
  dim double above/.style={to path={\pgfextra{
        \pgfinterruptpath
        \draw[>=latex,|->|] let
        \p1=($(\tikztostart)!3em!90:(\tikztotarget)$),
        \p2=($(\tikztotarget)!3em!-90:(\tikztostart)$)
        in(\p1) -- (\p2) node[pos=.5,sloped,above]{#1};
        \endpgfinterruptpath
      }
    }
  },
  dim below/.style={to path={\pgfextra{
        \pgfinterruptpath
        \draw[>=latex,|->|] let 
        \p1=($(\tikztostart)!-1em!-90:(\tikztotarget)$),
        \p2=($(\tikztotarget)!-1em!90:(\tikztostart)$)
        in (\p1) -- (\p2) node[pos=.5,sloped,below]{#1};
        \endpgfinterruptpath
      }
    }
  },
}

\tikzset{
    right angle quadrant/.code={
        \pgfmathsetmacro\quadranta{{1,1,-1,-1}[#1-1]}     
        \pgfmathsetmacro\quadrantb{{1,-1,-1,1}[#1-1]}},
    right angle quadrant=1, 
    right angle length/.code={\def\rightanglelength{#1}},   
    right angle length=2ex, 
    right angle symbol/.style n args={3}{
        insert path={
            let \p0 = ($(#1)!(#3)!(#2)$) in     
                let \p1 = ($(\p0)!\quadranta*\rightanglelength!(#3)$), 
                \p2 = ($(\p0)!\quadrantb*\rightanglelength!(#2)$) in 
                let \p3 = ($(\p1)+(\p2)-(\p0)$) in  
            (\p1) -- (\p3) -- (\p2)
        }
    }
}

\newcommand{\pgfextractangle}[3]{%
    \pgfmathanglebetweenpoints{\pgfpointanchor{#2}{center}}
                              {\pgfpointanchor{#3}{center}}
    \global\let#1\pgfmathresult  
}

\usetikzlibrary{shapes.arrows}
\newcommand{\commentarrow}[1][Azure4]{\tikz[baseline=-3pt]{\node[shape border uses incircle, fill=#1,rotate=180,single arrow, inner sep=1pt, minimum size=6pt, single arrow head extend=2pt]{};}}

\tikzset{ax/.style={-latex,line width=2pt}}

\tikzset{camera/.style={fill=Sienna1,fill opacity=0.5},%
image plane/.style={draw=RoyalBlue3,line width=2pt}}

\usetikzlibrary{graphs}
\newcommenter{todo}{Firebrick1}
\newcommenter{mmitjans}{Magenta}
\newcommenter{style}{Olive}

\newcommand{\pdv}[2]{\dfrac{\partial #1}{\partial #2}}
\renewcommand{\bf}[1]{\mathbf{#1}}
\newcommand{\generalvar}{\textcolor{black}{\bf{X}}}

\newcommand{\black}[1]{\textcolor{black}{#1}}
\renewcommand{\it}[1]{\emph{#1}}
\newcommand{\rarrow}[0]{$\rightarrow$ }

\title{\LARGE \textbf{
Koopman pose predictions for temporally consistent human walking estimations}
}

\author{Marc Mitjans$^{1}$, David M. Levine$^{2}$, Louis N. Awad$^{3}$, Roberto Tron$^{1}$%
\thanks{${^{1}}$Department of Mechanical Engineering, Boston University, 110 Cummington Mall, MA 02215, USA \{mmitjans, tron\}@bu.edu. Affiliates are supported by NIH R01AG067394-02. R. Tron is additionally supported by NSF NRI-1734454.}%
\thanks{$^{2}$Division of General Internal Medicine and Primary Care, Brigham and Women's Hospital, Harvard Medical School, 75 Francis St, Boston, MA 02115, USA \{dmlevine\}@bwh.harvard.edu.}
\thanks{${^{3}}$College of Health and Rehabilitation Sciences: Sargent College, Boston University, 635 Commonwealth, MA 02215, USA \{louawad\}@bu.edu. Affiliates are supported by AHA 18IPA34170487.}%
\thanks{We thank Michail Theofanidis, Aaron Horowitz and Jialan Sun for their invaluable help in labeling our datasets.}
}

\begin{document}

\maketitle
\thispagestyle{empty}
\pagestyle{empty}

\begin{abstract}
We tackle the problem of tracking the human lower body as an initial step toward an automatic motion assessment system for clinical mobility evaluation, using a multi-modal system that combines Inertial Measurement Unit (IMU) data, RGB images, and point cloud depth measurements. This system applies the factor graph representation to an optimization problem that provides \mbox{3-D} skeleton joint estimations.
In this paper, we focus on improving the temporal consistency of the estimated human trajectories to greatly extend the range of operability of the depth sensor.
More specifically, we introduce a new factor graph factor based on Koopman theory
that embeds the nonlinear dynamics of several lower-limb movement activities. This factor performs a two-step process: first, a custom activity recognition module based on spatial temporal graph convolutional networks recognizes the walking activity; then, a Koopman pose prediction of the subsequent skeleton is used as an \it{a priori} estimation to drive the optimization problem toward more consistent results.
We tested the performance of this module a dataset composed of multiple clinical lower-limb mobility tests, and we show that our \black{approach} reduces outliers on the skeleton form by almost 1 m, while preserving natural walking trajectories at depths \black{up to more than 10 m.}


\end{abstract}

\section{INTRODUCTION}

It is undeniable that the process of aging comes with an increasing frailty and loss of mobility. However, although
frail adults constitute the sickest, fastest growing, and most expensive segment of the US population \cite{medicare}, the US healthcare system still does not offer solutions to promptly detect the onset of frailty and act accordingly. Instead, it passively reacts to accidents that occur as a consequence of functional decline, which entail expensive and time-consuming hospital-based interventions. In this light, home-based therapies that monitor the motor skills of the patient on a continuous daily basis could allow therapists to make timely diagnoses of mobility decline, which would result in an increase of the patient's quality of life.
For this purpose, we propose an improved low-cost multi-modal system for human motion tracking, aimed at facilitating mobility assessment in the home through various every-day activities.
We build upon a preliminary version of the system \cite{mitjans2021visual}; such system uses a factor graph formulation to combine inertial measurements from four Inertial Measurement Units (IMUs, attached to the lower limbs of the user) with RGB and depth images from an depth camera, to provide \mbox{3-D} human movement estimations.
In this paper, we expand the factor graph formulation with
a novel factor that combines deep learning and Koopman theory to first predict the walking activity being performed, and then
compute a prediction of the user motion to bias the factor graph optimization toward a stronger temporal consistency.

\myparagraph{Related work}
Our work on human pose estimation is derived from a fusion between recent techniques used for robotic localization, namely Visual-Inertial Odomoetry (VIO), and data-driven machine learning models.
VIO techniques use a fusion of inertial measurements with visual information. There is also research in the literature on the use of IMUs, kinematic models and vision techniques for human pose prediction.
For instance, the authors in \cite{bai2014quantitative} used IMU data and a kinematic model to track the upper limbs in neurorehabilitation exercises; the work from
\cite{tian2015upper} showed that fusing IMU and vision data outperforms IMU-only approaches; and multiple studies, such as \cite{von2016human} and \cite{patil2021open}, have developed \mbox{3-D} pose estimation algorithms with IMU, LiDAR and multiple-camera data.

With the recent rise of deep neural networks, there has also been an increasing interest in tracking the human pose and recognizing human activities from a deep learning perspective.
For example, a common open-source library used for pose estimation is OpenPose \cite{cao2018openpose}, a Convolutional Neural Network (CNN)-based system for \mbox{2-D} multi-human pose estimation from RGB images. The authors of \cite{raaj2019efficient} proposed recurrent-based CNNs to track the human body with spatial and temporal information on video sequences; the authors of \cite{kocabas2019self}, \cite{nakano2020evaluation} and \cite{zhou2017towards} suggested methods to infer the \mbox{3-D} human joint positions from multiple \mbox{2-D} images; and the research from \cite{zhang2021cross} uses the Koopman operator to constrain the training of a deep autoencoder on silhouette images, whose feature space is then used for gait recognition.
More directly related to the problem of human activity recognition, state-of-the-art results are based on Graph Neural Networks (GCNs), which treat the human skeleton joints as graph nodes, and their connections (bones) as graph edges. For instance, the authors in \cite{heidari2021temporal} proposed a module to select the most informative frames in a skeleton sequence and fuse it with a GCN module;
the authors in \cite{zhang2019graph} presented a GCN architecture that fuses information both from nodes and skeleton edges; and the authors in \cite{yan2018spatial} introduced the spatial temporal GCN, which applies graph convolutions on the spatial domain and regular convolutions on the temporal domain.

\myparagraph{Paper contributions}
Our system explores the capabilities of Koopman theory in the field of human pose estimation.
We show that Koopman-based predictions of the human pose placed within the factor graph optimization loop condition the skeleton estimations towards more consistent gait estimations. This allows the system to obtain more realistic motion estimates even in the presence of significant noise (e.g.,  far from the camera), thus overcoming a main limitation in \cite{mitjans2021visual}. We analyze the performance of our system on a dataset of clinically-relevant lower limb mobility tests.


The remainder of the paper is organized as follows:  Sec.~\ref{sec:preliminaries} presents the preliminary concepts required to understand Sec.~\ref{sec:koopman-factor}, which describes the implementation of our proposed Koopman factor. Sec.~\ref{sec:experimental-study} presents the conducted experiments and the discussion of the results, and finally Sec.~\ref{sec:conclusions} summarizes the conclusions and introduces potential future work.

\section{PRELIMINARIES}
\label{sec:preliminaries}

We begin by providing a brief overview of the optimization framework for human walking estimations presented in \cite{mitjans2021visual}.
Then, we introduce the two new elements that constitute the core of this paper: the Koopman operator and the spatial-temporal graph neural network for walking activity selection.


\subsection{Human kinematic model}
\label{sec:prelim-walking-model}
We model the human kinematics with a directed tree, \mbox{$H = (V,E)$}. Each node $i\in V$ represents a joint of the articulated human model, and is associated to \mbox{3-D} trajectories $\bf{n}^i_k \in\real{3}$ expressed in an inertial reference frame, where $k$ is a discrete time index associated to a keyframe time $t_k$. Likewise, $(i,j) \in E$ represents a link connecting joints $i$ and $j$, and is associated to a time-invariant length $l^{ij}\in\real{}$, and a time-dependent extrinsic rotation $\bf{R}^{ij}_k \in SO(3)$, which represents the transformation from the inertial frame to a frame rigidly affixed to the link. By convention, the $z$-axis of each rotation is placed along its corresponding link pointing from parent node $i$ to child node $j$. Given the 3-D coordinates of the root node $\bf{n}^{0}_k$, this parameterization enables reconstructing the full human skeleton with the following kinematic relation:
\begin{equation}
\bf{n}^j_k = \bf{n}^i_k + l^{ij} \bf{R}^{ij}_k \bf{e}_3,
\label{eq:kinematic-relation}
\end{equation}
where $\bf{e}_3 = \left[0\,0\,1\right]^\top$ corresponds to the standard $z$ Cartesian coordinate basis.
Our human skeleton tree is composed of $7$ joints (feet, knees, hips, and sternum) and $6$ links, with the sternum acting as the root node.

We define the state of our human skeleton at time $t_k$ by concatenating the root node position $\bf{n}^{0}_k$ with all the link rotations $\bf{R}^{ij}_k$ to form state $\cR_k$. Likewise, we concatenate all link lengths $l^{ij}$ to form the set of parameters $\cL$.

\subsection{Factor graphs} \label{sec:prelim-factor-graphs}
Unrolling equation (\ref{eq:kinematic-relation}) for all joints and links across a sliding window of \emph{K} keyframes, we obtain a graph structure. This motivates representing our gait estimation problem using the factor graph formulation.
A factor graph~\cite{kschischang2001factor} $G=(S,F)$ is a bipartite graph which represents a complex multivariate probabilistic model as a multiplication of simpler models (\emph{factors}). Each factor $f_k\in F$ for $k\in K$ depends on a small subset of the system variables (\emph{states}) $s_k\in S$ and encodes the likelihood of $s_k$ given certain external measurements $z_k\in Z$, $f_k(s_k;z_k)$. The joint probability $P(S|Z)$ of all the states $S=\{s_k\}$ given all the measurements $Z=\{z_k\}$ can then be expressed as
\begin{equation}
    P(S|Z) \propto \prod_{k\in K} f_k(s_k; z_k).
    \label{eq:joint-probability}
\end{equation}

The maximum likelihood estimation (MLE) of $S$ is obtained by minimizing the negative log-likelihood of $P(S|Z)$. Assuming that factors $f_k$ follow zero-mean Gaussian distributions, one can express the MLE of (\ref{eq:joint-probability}) as the least-squares solution of an optimization problem depending on residuals $\bf{r}_k(s_k)\in\real{n}$ \cite{loeliger2004introduction},
\begin{equation}
    S^* = \arg\min -\ln P(S|Z) = \arg\min -\sum_{k\in K} \|\bf{r}_k\|^2_{\Sigma_k},
    \label{eq:optimizatiopn-prob}
\end{equation}
where $\Sigma_k$ corresponds to the noise covariance of the measurement $z_k$ associated to the factor $f_k$. 
\color{black} The residuals $\bf{r}_k$ express the difference between the measurement $z_k$ and the (possibly nonlinear) measurement function $h_k(s_k)$.
\color{black}
This representation of the factor graph optimization problem intuitively defines $\Sigma^{-1}_k$ as the \emph{weight} associated to each residual $\bf{r}_k$. With this insight, we can use factors that are based on machine learning algorithms, treating their weights (covariance matrices) as design parameters.

The optimization problem \eqref{eq:optimizatiopn-prob} is generally solved using gradient descent or quasi-Newton methods; this requires computing the gradients of the residuals, which requires finding expressions for the Jacobian algorithms with respect to the optimization variables.


\subsection{Visual-inertial 3-D human pose estimations}
\label{sec:visual-inertial-introduction}

We previously presented a visual-inertial system for reconstructing \mbox{3-D} human poses~\cite{mitjans2021visual}; this system applies the factor graph formulation described in Sec.~\ref{sec:prelim-factor-graphs} to the kinematic model from Sec.~\ref{sec:prelim-walking-model} to estimate the 3-D coordinates of the root node $\bf{n}^{0}_k$, the extrinsic rotations $\bf{R}^{ij}_k$, and the intrinsic link lengths $l^{ij}$, over a sliding window to approximate the \mbox{3-D} skeleton movement.
This is achieved by fusing information from four IMUs strapped on the two legs (two shanks and two thighs) with synchronized images and depth point clouds captured by an RGB-D camera. RGB images provide \mbox{2-D} pixel coordinates of the human skeleton joints, while the point cloud provides their depth component. 
\color{black}
These measurement sources are encoded in four different types of factors over their corresponding subset of state variables: \begin{enumerate*}
    \item The IMU factor, which computes the estimated rotation change of the links between adjacent keyframes by applying \emph{preintegration} theory \cite{forster2015imu} on the angular velocities;
    \item the Image factor, which computes the reprojection error between the $xy$ coordinates of the predicted joints under perspective projection~\cite{hartley2003multiple} and the pixel coordinates measured by OpenPose;
    \item the Depth factor, which computes the depth error on the estimated joints; and
    \item the Contact factor, which uses a logistic regressor on IMU data to predict feet contacts and thus avoid estimating unrealistic skids (sliding of the feet).
\end{enumerate*}





\color{black}

This preliminary work was implemented using the GTSAM library for factor graph optimization~\cite{dellaert2012factor}, and it
showed the potential of visual-inertial filtering and factor graphs applied to 3-D human pose estimation. However, the current technology for time-of-flight cameras leads to noise variances that increase with the distance from the sensor; this fact considerably affected the depth range of operability of the system in \cite{mitjans2021visual}, highly decreasing its performance beyond distances further than \unit[$7$]{m} away from the camera.


\subsection{Koopman theory}
\label{sec:koopman-introduction}

The time evolution of the \mbox{3-D} skeleton joint coordinates estimated in Sec.~\ref{sec:visual-inertial-introduction} during a certain walking activity can also be regarded as an autonomous nonlinear dynamical system,
\begin{equation}
    \bf{x}_{k+1}= f(\bf{x}_k),
    \label{eq:general-dynamical-system}
\end{equation}
where $\bf{x} \in \real{21}$ (7 joints with 3 coordinates each), and $f$ is the nonlinear evolution operator. This description enables the exploration of additional techniques for nonlinear systems to predict more accurate joint states.
For this purpose, here we introduce a general overview of the discrete-time Koopman operator. We refer the interested reader to \cite{williams2015data} for a more in-depth development.

Let $\bf{x}$ be an \emph{n}-dimensional state variable on a manifold $\cM \subseteq \real{n}$, whose discrete-time evolution is defined by (\ref{eq:general-dynamical-system}),
and where \mbox{$f : \cM \mapsto\cM$} is the (nonlinear) evolution operator. Moreover, let us define an \emph{observable} function $\psi$ that belongs to the set of functions mapping elements of $\cM$ to $\real{}$, $\cF = \{f: \cM\mapsto\real{}\}$. Then, the Koopman operator $\cK$ is defined as an operator acting on $\psi$ such that
\begin{equation}
    (\cK\psi)(\bf{x}) = (\psi \circ f)(\bf{x}) = \psi(\bf{x}_{k+1}).
    \label{eq:koopman-eq}
\end{equation}

Equation (\ref{eq:koopman-eq}) describes a new dynamical system in $\cF$, where $\cK$ determines the evolution dynamics of the observable function $\psi$, contrary to $f$ which acts directly on $\cM$ \color{black}($\circ$ is the composition operator)\color{black}. Additionally, this definition of the Koopman operator holds two interesting properties.
On the one hand, since $\cF$ is infinite dimensional, then also $\cK$ must be infinite dimensional, thus rendering this exact definition impractical for real-life applications. On the other hand, $\cF$ is a vector-valued space and the composition operator is linear on this space, and hence the Koopman operator is also linear, even if the initial dynamics governing $f$ are nonlinear. The data-driven Extended Dynamic Mode Decomposition (EDMD) \cite{williams2015data} method approximates the infinite-dimensional Koompan operator $\cK$ by a finite-dimensional operator $\bf{K}$ (a matrix), which in turn yields a linear approximate representation of the initial system.

EDMD requires a training dataset $D$ with $M$ consecutive pairs of system states, $D=\{(\bf{x}_i,f(\bf{x}_i)\}_{i=0}^{M-1}$. Additionally, let us define a finite set of $P$ observable functions that form the column vector-valued function $\bf{\Psi} : \cM \mapsto \real{P}$ (which are typically chosen by the designer, see Sec.~\ref{sec:koopman observables}),
\begin{equation}
    \bf{\Psi}(\bf{x}) = \left[\psi_1(\bf{x}),\psi_2(\bf{x}),\dots,\psi_P(\bf{x})\right]^\top.
    \label{eq:Psi}
\end{equation}

By combining equations (\ref{eq:koopman-eq}) and (\ref{eq:Psi}) we arrive at the following expression, which is linear in $\bf{\Psi}$:
\begin{equation}
    \bf{\Psi}(\bf{x}_{k+1}) \approx \bf{K} \bf{\Psi}(\bf{x}_k).
    \label{eq:final-koopman}
\end{equation}

The approximate Koopman matrix $\bf{K}$ can be computed from (\ref{eq:final-koopman}) in closed form
using all data points in $D$ by solving its corresponding least-squares problem \cite{williams2015data}.

\begin{remark}[Predicting $\bf{x}_{k+1}$] \label{remark:predicting-x}
The system state $\bf{x}_{k+1}$ can be predicted using the approximate Koopman operator by using the identity map as one of the vector-valued observable functions \cite{abraham2017model}, namely
%
\begin{equation}
    \bf{\Psi}(\bf{x}) = [\bf{x}^\top,\psi_1(\bf{x}),\psi_2(\bf{x}),\dots,\psi_P(\bf{x})]^\top.
\end{equation}
The approximate prediction $\bf{\hat{x}}_{k+1}$ is then extracted with
\begin{equation}
    \bf{\hat{x}}_{k+1} = \bf{K}^{[n]} \bf{\Psi}(\bf{x}_k) \approx \bf{x}_{k+1},
\end{equation}
where $\bf{K}^{[n]}$ corresponds to the first $n$ rows of the Koopman matrix $\bf{K}$. Note that the extended $\bf{\Psi}(\bf{x})$ needs to be defined prior to computing $\bf{K}$.

\end{remark}

\subsection{Spatial Temporal Graph Convolutional Network}

Sec.~\ref{sec:koopman-introduction} considers a way to create a model of human movement as an autonomous dynamical system. This model might be reasonable when a single activity is considered (e.g., walking). However, real-life data is likely to comprise a sequence of multiple activities, such as \emph{standing up} \rarrow \emph{standing} \rarrow \emph{walking}. In this case, we posit that it is more natural to model the full sequence as the evolution of a hybrid dynamical system with three underlying and fundamentally different models. We therefore propose to use a classifier that can distinguish between activities to select the right model.
In addition, the graph-like structure of the human skeleton and the correlations between the motions suggest utilizing an architecture that can capture this organization. For this purpose, we use a Spatial Temporal Graph Convolutional Network~\cite{yan2018spatial} architecture (ST-GCN), which we review below.

An ST-GCN is a neural network that receives both the joint coordinates from a sequence of keyframes and information on the graph structure, and performs a series of spatio-temporal (ST) convolutions to extract features both in the spatial domain and the temporal domain. Each ST layer implements one convolution of each type.

\subsubsection*{Spatial Convolution}
This convolution exploits the natural intra-skeleton connections (edges) between the joints on a single keyframe, and is defined as
\begin{equation}
    f_{out}(v_{i}) = \sum_{v_{j}\in B(v_{i})} \dfrac{1}{Z_{ij}} f_{in}(v_{j}) w(l_{i}(v_{j})).
    \label{eq:spatial-convolution}
\end{equation}

$f_{in}$ and $f_{out}$ are the input and output features for a given node, respectively, $v_i$ is the center 
node where the convolution is computed,
$B(v_i)$ corresponds to the set of 1-distance \emph{spatial neighbors} of node $v_i$, and ${w}$ is the learnable weight variable. While this expression resembles the standard \mbox{2-D} image convolution, the main difference lies on the partitioning strategy used to divide the neighbors into subsets, governed by the mapping function $l_i$.
In particular, $l_i(v_j)$ maps each $v_j \in B(v_i)$ to one of 3 different subsets: the center joint $v_i$, the centripetal group (neighbors closer to the centroid of the skeleton than $v_i$), or the centrifugal group (neighbors further from the center than $v_i$). Then, $Z_{ij}$ becomes a scaling factor that accounts for unbalanced neighbor contributions to each group.

The spatial convolution from equation (\ref{eq:spatial-convolution}) can be implemented in tensor form \cite{kipf2016semi} for each frame $t$ in $T$ as
\begin{equation}\label{eq:st gcn}
    \bf{F}_{out,t} = \sum_{j=1}^3 \Lambda_{jt}^{-\frac{1}{2}} (\bf{A}_{jt} \odot \bf{M}) \Lambda_{jt}^{-\frac{1}{2}} \bf{F}_{in,t} \bf{W}_j.
\end{equation}

$\bf{F}_{in,t}$ and $\bf{F}_{out,t}$ are slices on the temporal dimension of the input and output tensors, of shape $\left[N,T,C_{in}\right]$ and $\left[N,T,C_{out}\right]$ respectively ($N$ corresponds to the number of joints, $T$ denotes the temporal dimension, $C$ is the number of channels), and $\bf{W}_j$ represents the weight tensor for group $j$, of shape $[C_{in},C_{out}]$.
The structure of the graph connectivity is encoded by the $N\times N$ matrices $\bf{A}_{jt}$, which represent the adjacency matrices of all one-hop neighbors on each subset; $\Lambda_{jt}$, which are diagonal matrices
whose diagonal elements are $\Lambda^{ii}_{jt} = \sum_{m=0}^{N-1} \bf{A}^{im}_{jt} + \alpha$ (with $\alpha \ll 1$ to avoid singularities);
and $\bf{M}_j$, which are mask matrices that learn the importance of each edge connection ($\odot$ denotes the element-wise product). 

\subsubsection*{Temporal Convolution} The temporal convolution is applied on the output tensor of the spatial convolution phase. It exploits the inter-frame connections of the same joint between adjacent keyframes by performing a standard \mbox{1-D} convolution on the dimension $T$ of $\bf{F}_{out}$. Hence, in this case the \emph{temporal neighbors} of node $v_{it}$ are defined as \mbox{$B(v_{it}) = \{v_{iq} | -\Gamma/2 \leq q-t \leq \Gamma/2 \}$}, where $\Gamma$ is the temporal kernel size.

Lastly, the output of the final ST layer is sent to a fully connected (FC) layer, which predicts the activity label.



\section{THE KOOPMAN FACTOR}
\label{sec:koopman-factor}

\begin{figure}[t]
\centerline{\includegraphics[width=0.8\columnwidth]{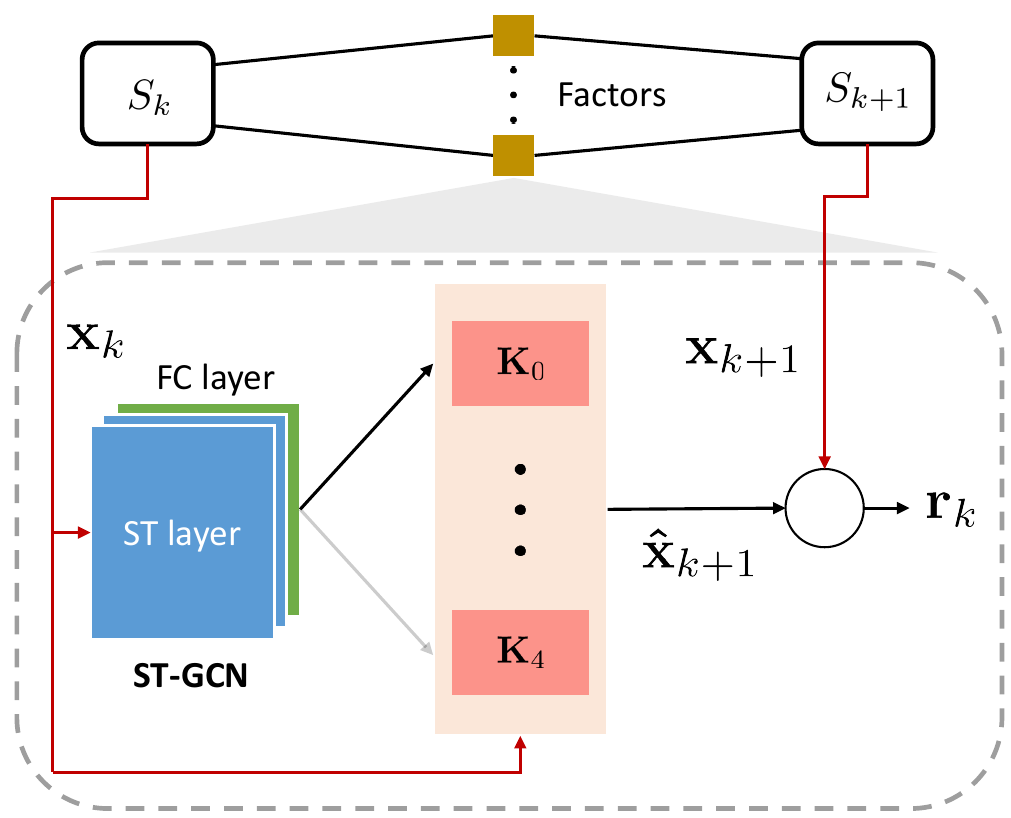}}
\caption{The schematic of the Koopman factor on a two-state window. The state at $\cS_k$ is used to compute the joint coordinates $\bf{x}_k$, which are sent to the activity selector. The selected Koopman matrix predicts the next joint states, which in turn are used to compute the residual of the factor.
}
\label{fig:koopman-factor}
\end{figure}

In this paper, our goal is to build a new factor graph factor that, given a set of estimated joint coordinates \mbox{$\bf{x}_k = [\bf{n}^{0}_k,\dots,\bf{n}^6_k]^\top$} at time $t_k$, it
\begin{enumerate*}
\item identifies its corresponding walking activity $a_k$, and
\item predicts a \emph{prior} new set of joints at $t_{k+1}$ by selecting a Koopman matrix $\bf{K}_{a_k}$ associated to activity $a_k$.
\end{enumerate*}
This factor is added to the original factor graph from Sec.~\ref{sec:visual-inertial-introduction}, and the  schematic of its components and architecture is shown in \black{Fig.~\ref{fig:koopman-factor}}.

\subsection{Normalization of input variables}

The two trainable modules receive the \mbox{3-D} skeleton joint coordinates as inputs. The map between the system variables $\left\{\cR_k,\cL\right\}$ and a joint $\bf{n}^j_k$ is defined as

\begin{equation}
    \label{eq:ori_integr_unravel}
    \bf{n}^j_k = \bf{n}^0_k +
    \sum_{(p,q)\in E_j} l^{pq} \bf{R}_k^{pq} \bf{e}_3,
\end{equation}
where $(p,q)$ runs across all links $E_j$ in the kinematic chain from the root node $\bf{n}^0$ to the child node $\bf{n}^j$.

To make the system invariant to translation and scaling, the input skeletons to both systems are previously centered around their right foot (to preserve as much skeleton variance information as possible \cite{vox2018preprocessing}) and scaled down to be within the range $[0,1]$. We denote this centering and normalization operation as $\cN$, and it is applied to each individual joint $\bf{n}^j$:
\begin{equation}
    \bf{\bar{n}}^j_k = \cN (\bf{n}^j_k) = \dfrac{\left(\bf{n}^j_k - \bf{n}^{(3)}_k\right) - c_{min}}{c_{max} - c_{min}}.
    \label{eq:n-map}
\end{equation}

$\bf{n}^{(3)}_k$ are the coordinates of the 3rd joint, which corresponds to the right foot, and are used for centering the skeleton.
$c_{min}$ and $c_{max}$ are scalars that refer to the minimum and maximum coordinate values over the whole training dataset. These values are stored in memory for the normalization of other datasets. Note that
the vector $\bf{x}_k$ can be built by concatenating all seven $\bf{n}^j_k$.





\subsection{ST-GCN for per-frame walking activity selection}
\label{sec:intro-stgcn}

Given a temporal sequence of skeleton joint coordinates, we seek to build a classifier that determines the \emph{walking activity type} for each individual keyframe $k$.
For this study we divided the walking activities into five different labels $a$: \emph{walking}, \emph{standing}, \emph{sitting}, \emph{standing up}, and \emph{sitting down}, such that $a \in \left[0,\dots,4\right]$. These categories fully encompass all the activities performed by the participants across all trajectories studied in Sec.~\ref{sec:experimental-study}.



Many of the GCN-based activity recognition studies in the literature focus on predicting a single activity label from a complete video sequence. In these cases, the temporal dimension usually comprises the entirety of each trajectory.
However, our specific application demands \emph{per-frame} labeling. This fact highly constrains the temporal dimension size $T$, and thus requires splitting each full sequence into multiple independent subset trajectories of a much smaller $T$ size. This restriction also has a direct impact on the selection of both the temporal kernel size $\Gamma$ and the number of ST layers.
Regarding the spatial convolution, the raw data from the joint trajectories is used to pre-compute the tensors $\bf{A}_j$ and $\Lambda_j$ taking into account the link connections in the skeleton structure.


\subsection{Koopman observable functions}\label{sec:koopman observables}

One core element in the construction of the Koopman matrix is the choice of the basis functions to form the vector of observables $\bf{\Psi}$. Many different strategies can be followed, which highly depend on the application in hand. However, there is a lack of literature on the application of Koopman theory to human gait estimations due to its novelty. For this reason, we empirically selected the univariate Fourier basis \cite{konidaris2011value} due to its simplicity and good performance, and encourage future research to study the impact the choice of basis functions has on the quality of the gait estimations.

Let $\bf{x}\in\real{d}$. Its univariate $n$th-order Fourier basis is defined as the set of functions such that
\begin{equation}
    \psi_n(\bf{x}) = \cos(\pi \bf{c}_n^\top \bf{x}),
\end{equation}
where $\bf{c}_n = [c_0,\dots,c_{d-1}]^\top$ is a coefficient vector of size $d$, and $c_j \in [0,\dots,n]$ for $c_j \in \bf{c}$. This implies that for the $d$-dimensional variable $\bf{x}$, its Fourier basis is composed of $(n+1)^d$ basis functions, a number that suffers from the curse of dimensionality.
Since in our application \mbox{$\bf{x}\in\real{21}$}, this representation becomes initially impractical. In this exploratory study, we decouple the three dimensions $xyz$ of the joint coordinates and feed them to $\bf{K}$ independently, which allows reducing the feature dimension from $21$ to $7$. This heuristic approach \black{yielded} mean test errors of less than \unit[$3$]{cm} per frame. Empirically, for this application we selected $n=1$, and applying Remark \ref{remark:predicting-x} it renders Koopman matrices \mbox{$\bf{K} \in\real{135\times135}$}.


\subsection{The Koopman factor}
\label{sec:the-koopman-factor}

\begin{figure}[h]
\centerline{\includegraphics[width=\columnwidth]{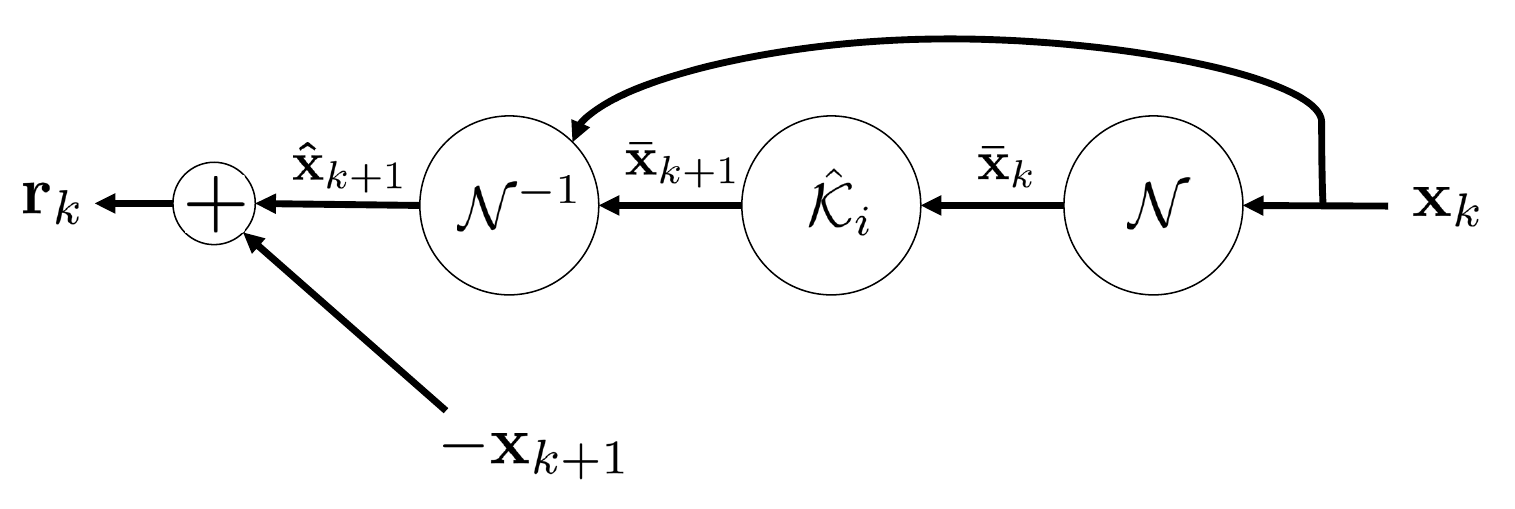}}
\caption{The computational graph with the operations performed inside the Koopman factor. $\cN$ and $\cN^{-1}$ are the normalization map and its inverse respectively, and $\hat{\cK}_i$ is our Koopman operator applied to coordinate $i$ of $\bf{\bar{x}}_k$.
}
\label{fig:computational-graph}
\end{figure}

Given two consecutive gait state estimations, $\cS_k= \left\{ \cR_k,\cL\right\}$ and $\cS_{k+1} = \left\{ \cR_{k+1},\cL\right\}$, the Koopman factor performs the
following operations sequentially (Fig.~\ref{fig:computational-graph}):
\begin{enumerate}
    \item Obtains the joint coordinates $\bf{x}_k$ and $\bf{x}_{k+1}$ from the factor graph estimation,
    \item normalizes $\bf{x}_k$ by applying equation (\ref{eq:n-map}),
    \item computes the observable vectors $\bf{\Psi}_x(\bf{\bar{x}}_k)$, $\bf{\Psi}_y(\bf{\bar{x}}_k)$, $\bf{\Psi}_z(\bf{\bar{x}}_k)$ associated to each coordinate of $\bf{\bar{x}}_k$,
    \item predicts activity $a_k(\bf{\bar{x}}_k)$ and selects  $\bf{K}_{a_k}$,
    \item predicts the coordinates of the joint states $\bf{\hat{x}}_{k+1}$, and
    \item computes the residual $\bf{r}_k = \bf{\hat{x}}_{k+1} - \bf{x}_{k+1} \in\real{21}$.
\end{enumerate}

$\hat{\cK}_i = \bf{K}^{[7]}_{a_k} \bf{\Psi}_i$ is our Koopman map applied on coordinate $i\in\{x,y,z\}$,
$\cN^{-1}$ is the inverse of equation (\ref{eq:n-map}),
and $\bf{\bar{x}}_{k+1}$ is obtained by stacking the three coordinate column vectors in the first dimension,
\mbox{$\bf{\bar{x}}_{k+1} = [ \left(\bf{\bar{x}}^\top_{k+1}\right)_x,
\left(\bf{\bar{x}}^\top_{k+1}\right)_y, \left(\bf{\bar{x}}^\top_{k+1}\right)_z]^\top$}.
Note that to make the centering of the skeleton consistent, the right foot coordinates $\bf{n}^{(3)}_k$ are also used in $\cN^{-1}$ to recover $\bf{\hat{x}}_{k+1}$.

\myparagraph{Jacobians}
As required by most optimization-on-manifold solvers, explicit expressions for Jacobians of the residual with respect to each variable are required.
In the remainder of this section, we provide expressions for the Jacobians of $\bf{r}_k$ for coordinate $i$ of an arbitrary node $\bf{n}^j_k$ with respect to a parent link $(p,q)$. We also drop the sub-index $k$ for an easier readability.
By applying the chain rule, we can express the Jacobian of $\bf{r}^j$ with respect to a general variable $\generalvar$ as
%
\begin{equation}
    \pdv{\bf{r}^j_i}{\generalvar} =
    \pdv{\cN^{-1}}{\hat{\cK}_i}
    \pdv{\hat{\cK}_i}{\cN}
    \pdv{\cN}{\bf{X}} +
    \pdv{\bf{n}^{(3)}_i}{\bf{X}}
    - \pdv{(\bf{x}_{k+1})_i}{\bf{X}}.
    \label{eq:jacobian-wrt-general}
\end{equation}

We provide the mathematical expressions for each term in (\ref{eq:jacobian-wrt-general}) below:
\begin{subequations}
\begin{align}
    \pdv{\cN^{-1}}{\hat{\cK}_i} &= (c_{max} - c_{min}), \\\
    \pdv{\hat{\cK}_i}{\cN} &= \bf{K}_{a_k}^{[7]}
    \left[ \bf{I}_7, \left[
    -\sin \left( \pi \bf{c}_m^\top \bf{\bar{x}}_i  \right) \pi \bf{c}_m
    \right]_{\bf{c}_m\in\cC} \right]^\top, \\
    \pdv{\cN}{\bf{X}} &=
    %
    \dfrac{1}{c_{max}-c_{min}} \left( \pdv{\bf{x}_i}{\generalvar} - \pdv{\bf{n}^{(3)}_i}{\generalvar}  \right).  \label{eq:der-x-hat}
\end{align}
\end{subequations}

$\bf{x}_i$ is a \mbox{7-D} vector containing coordinate $i$ for all joints, $\bf{I}_7$ corresponds to the \mbox{7-D} identity matrix,
and $\cC$ corresponds to the set of all frequencies $\bf{c}_m$ included in $\bf{\Psi}_i$.
The task to compute the Jacobians with respect to each state variable is now reduced to providing individual expressions for the Jacobians of each joint coordinate, analyzed below.

\subsubsection{Jacobian with respect to $\cR_k$}

Recall that the set of states $\cR_k$ is defined as the concatenation of all $\bf{R}^{ij}_k$ for $(i,j)\in E$ and $\bf{n}^0$.
The derivatives of a joint $\bf{n}^j$ with respect to a parent $\bf{R}^{pq}$ and the root joint $\bf{n}^0$ can be expressed as
\begin{subequations}
\begin{align}
    \pdv{\bf{n}^j}{\bf{R}^{pq}} &= \left[ -\bf{R}^{pq} \bf{e}_2, \bf{R}^{pq} \bf{e}_1, \bf{0}_{3\times1} \right] l^{pq},
    \label{eq:derivatives-joint-R} \\
    \pdv{\bf{n}^j}{\bf{n}^0} &= \bf{I}_3.
\end{align}
\label{eq:jac-r}
\end{subequations}
%


\subsubsection{Jacobian with respect to $\cL$}

$\cL$ refers to the concatenation of all link lengths $l^{ij}$ for $(i,j)\in E$. The derivative of $\bf{n}^j$ with respect to a parent $l^{pq}$ is
\begin{equation}
    \pdv{\bf{n}^j}{l^{pq}} = \bf{R}^{pq} \bf{e}_3.
    \label{eq:jac-links}
\end{equation}

Equations (\ref{eq:jac-r}) and (\ref{eq:jac-links}) should be computed for each joint and rearranged to form three \mbox{7-D} column vectors, corresponding to each coordinate $x$, $y$ and $z$, before plugging them into (\ref{eq:jacobian-wrt-general}).


\section{EXPERIMENTAL STUDY}
\label{sec:experimental-study}

\subsection{Dataset}
\label{sec:training-dataset}

The dataset used in this study
comprises 65 trajectories across 5 participants corresponding to four different standardized lower limb mobility tests, namely the 10 Meter Walk Test (10MWT), the Functional Gait Assessment (FGA), the Short Physical Performance Battery test (SPPB), and the Timed Up and Go test (TUG). Each trajectory contains ground truth data from the Qualisys motion capture system at \unit[200]{Hz}, RGB and point cloud data at \unit[30]{Hz} from an Intel RealSense D435~\cite{real}, and IMU data at \unit[120]{Hz}. This dataset was recorded under the IRB protocol Mass General Brigham 2020P003474, and the data was managed by the Robotic Operating System (ROS)~\cite{ros} middleware.

To build a joint trajectory dataset from the recorded data, we first applied the original system in \cite{mitjans2021visual} to each individual recording, which resulted in 65 different joint trajectories, each one comprising approximately between 150 to 600 keyframes (depending on the activity performed).
These trajectories contained skeleton keyframes up to \unit[$7$]{m} away from the camera, to remain in the operability range of the system.
To make the systems rotation invariance, each skeleton keyframe belonging to the training dataset was duplicated at different rotation angles around the vertical axis after the centering operation.

\subsection{ST-GCN training}

Considering the requirements addressed in Sec.~\ref{sec:intro-stgcn}, our ST-GCN is composed of two spatio-temporal layers with a temporal kernel size of $\Gamma=7$, with \black{128} and \black{256} output channels respectively, and a fully connected layer. This architecture sets the input tensor to be of shape $[B,C_{in},N,T]$ with $C_{in}=3$ spatial coordinates ($x$, $y$ and $z$), $N=7$ joints, and $T=13$ keyframes. $B$ corresponds to the batch dimension.
After centering each skeleton, we duplicated each keyframe 7 times while rotating the skeleton at intervals of $45^\circ$ around the vertical axis.
Then we split it into train and test stratified sets, resulting in $\black{66645}$ and $\black{28563}$ samples per set ($70\%$ and $30\%$ of the full dataset respectively). Each class on the train set has $13109$, $4609$, $3472$, $3041$ and $42414$ samples.
We implemented our ST-GCN using the Pytorch library \cite{NEURIPS2019_9015}.


To validate the choice of selecting joint coordinates as input features to our network, and
since our datasets already contain IMU information (unlike all the other cited activity recognition references), we compared our joint-based ST-GCN activity selector with an IMU-only ST-GCN baseline with a similar architecture.
This baseline network takes the pre-integrated IMU velocity measurements as inputs,
and the spatial subsets were generated by comparing the root node and its neighbors to the average speed of the four IMUs.


The joint-based ST-GCN output $99.97\%$ test accuracy, while the IMU-only baseline stayed at $92\%$. Additionally, many of its misclassification errors happened between the standing and sitting classes. This makes sense intuitively, as in both cases the IMU angular velocities are very close to zero. We leave for future work more thorough studies on the use of IMU-only data for activity recognition.

\subsection{Koopman training}

\begin{figure}
    \centering
    \subfloat[Standing]{
    \includegraphics[width=.32\columnwidth]{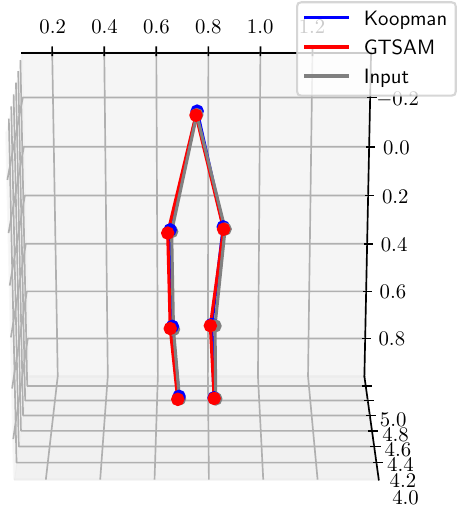}}
    \subfloat[Sitting]{
    \includegraphics[width=.32\columnwidth]{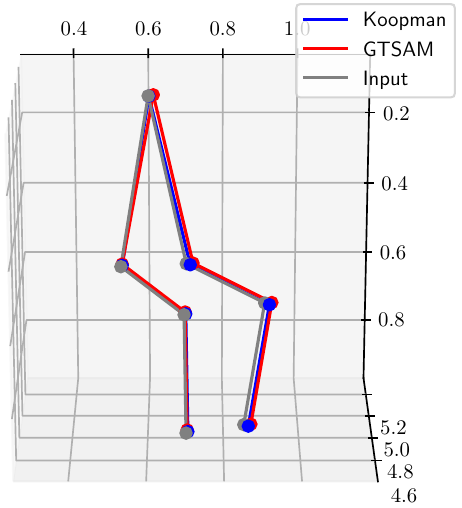}}
    \subfloat[Standing up]{
    \includegraphics[width=.32\columnwidth]{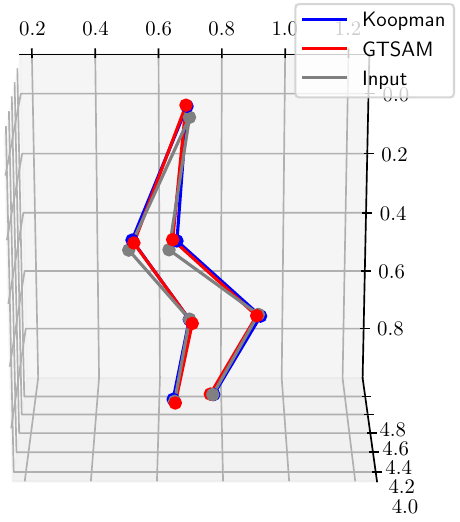}}
    \hfill
    \subfloat[Sitting down]{
    \includegraphics[width=.32\columnwidth]{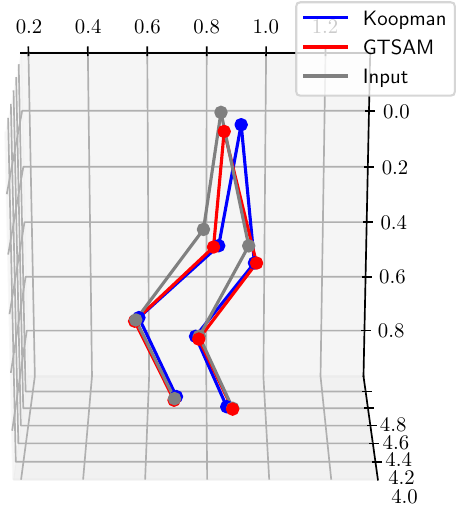}}
    \subfloat[Walking]{
    \includegraphics[width=.32\columnwidth]{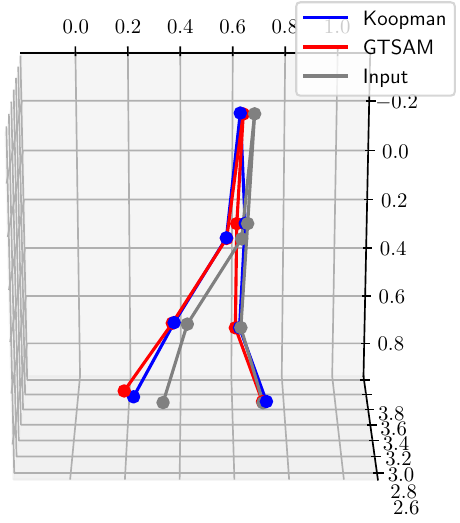}}
    \hfill
    \caption{Comparison between the Koopman predictions with their corresponding GTSAM estimations for one test sample of each one of the five selected walking activities. In gray, the GTSAM skeleton at $t_k$; in red, the GTSAM estimation at $t_{k+1}$; in blue, the Koopman prediction at $t_{k+1}$.
    \vspace{-5mm}
    }
    \label{fig:koopman-training}
\end{figure}

To construct the Koopman training dataset, all joint trajectories were segregated in pairs of consecutive keyframes $\{\mathbf{x}_k,\bf{y}_k\}$.
Contrary to the dataset used to train the ST-GCN, both keyframes were centered around the same right foot coordinates of $\bf{x}_k$ to match the computational graph from Fig.~\ref{fig:computational-graph}.
Then, both skeletons of each pair were duplicated 23 times and rotated at intervals of $15^\circ$ around their vertical axes.
Finally, the pairs were assigned an activity label $a_k$ corresponding to $\bf{x}_k$. This allows training one Koopman matrix $\bf{K}_{a_k}$ for each activity independently but following the same procedure.
The five Koopman matrices were trained following equation (\ref{eq:final-koopman}) and stored in memory.
Fig.~\ref{fig:koopman-training} shows five frame comparisons belonging to the test sets between the Koopman predictions and their corresponding GTSAM estimations.


\subsection{Experimental analysis}

We analyzed the performance of our newly proposed factor by comparing its results with a naive vision-only approach (consisting of image and depth data), with the GTSAM baseline system from \cite{mitjans2021visual}, and with their corresponding ground truth trajectories. 

\begin{figure}[h]
\centering
\includegraphics[trim={0 0mm 0 0},clip,width=0.8\columnwidth]{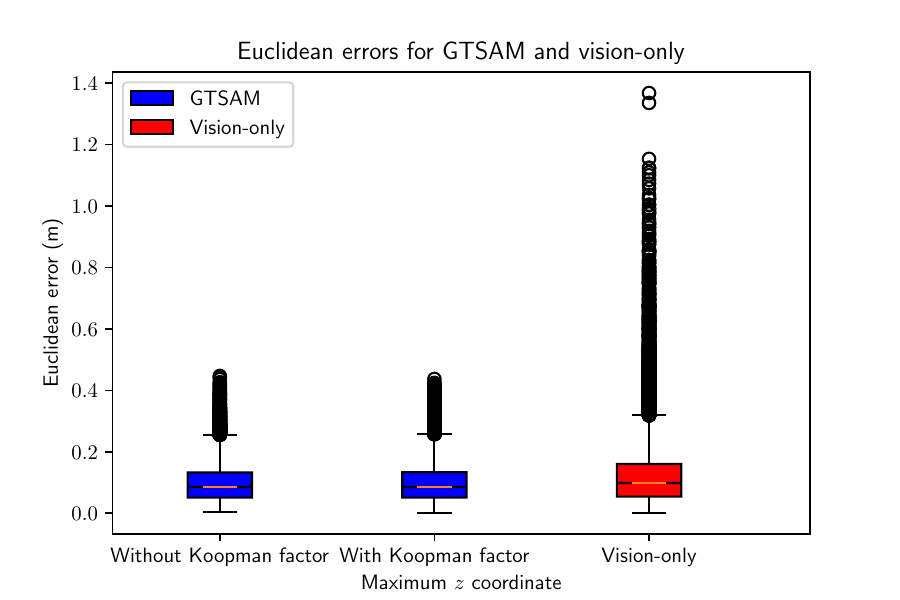}
\hfill
\caption{Boxplot comparison of the Euclidean errors of all joints across all trajectories with respect to their ground truth counterparts.
}
\label{fig:all-boxplots}
\end{figure}

To study the similarities between the estimated skeleton and its ground truth, we first centered the skeletons from each keyframe of every source around their centroid; then, we computed the Euclidean distance error with respect to the ground truth for all joints across all keyframes in the dataset. This approach allows removing the trajectory estimation bias while focusing solely on the skeleton form. \black{Fig.~\ref{fig:all-boxplots}} shows a boxplot comparison of all the Euclidean errors across all keyframes of all trajectories between the vision-only approach, the baseline system, and the modified factor graph. While both factor graphs highly improve the naive vision-only approach (they reduce the outliers by almost \unit[$1$]{m}), there is no considerable difference between the quality of the skeletons from both systems. This was to be expected, since the Koopman matrices were trained purely with the baseline estimations.

\begin{figure}[h]
\centering
\subfloat[Vision-only trajectories]{\includegraphics[width=0.8\columnwidth]{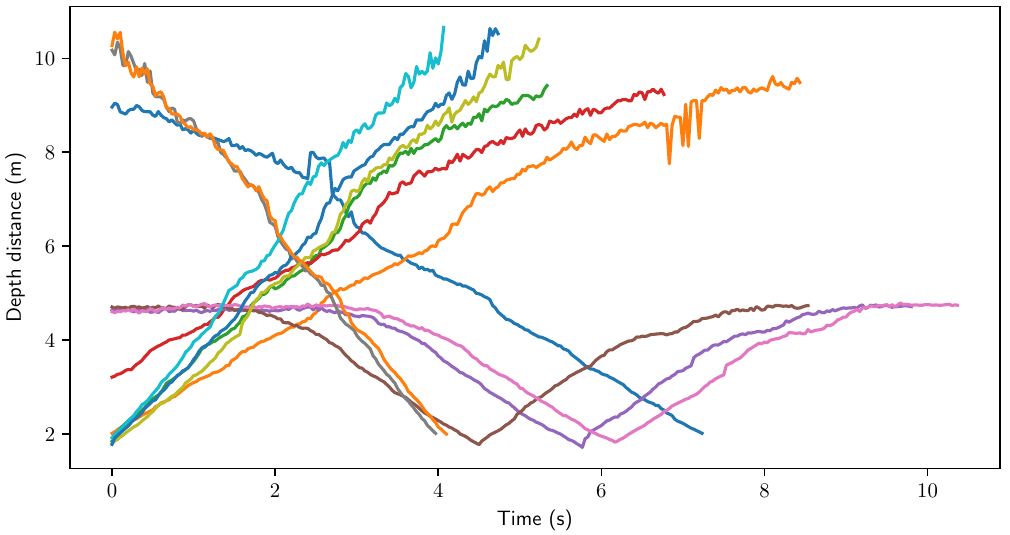}
\label{fig:vision-only-depth}
} \hfill
\subfloat[Without the Koopman factor]{\includegraphics[width=0.8\columnwidth]{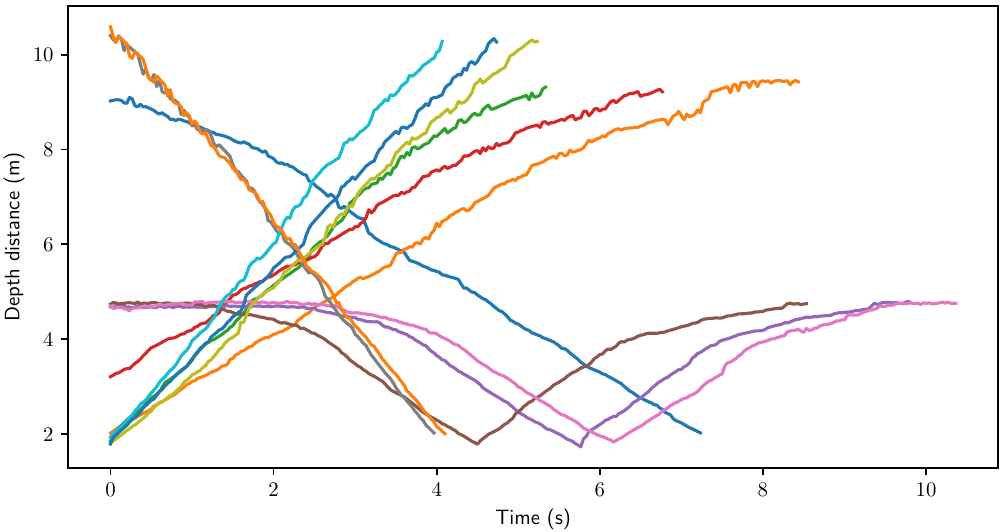}
\label{fig:old-GTSAM-depth}
} \hfill
\subfloat[With the Koopman factor]{\includegraphics[width=0.8\columnwidth]{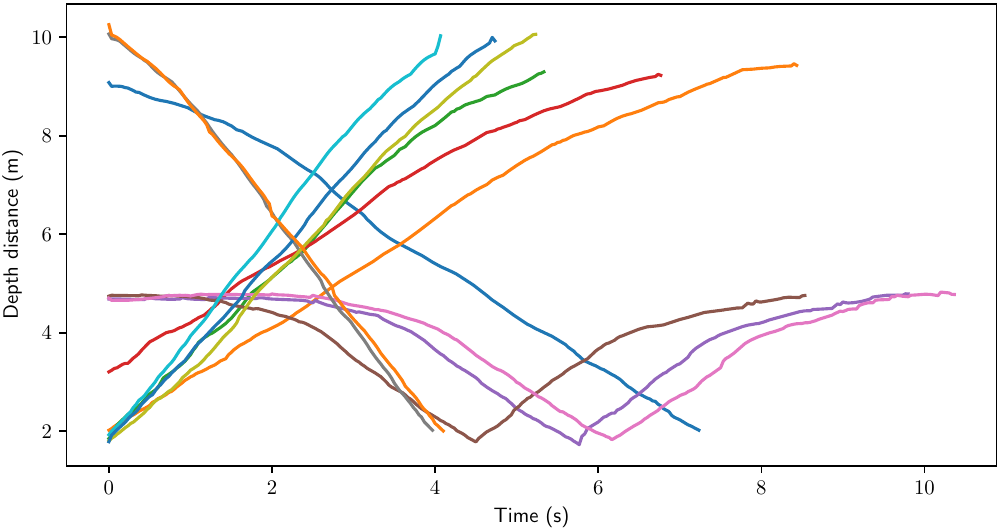}
\label{fig:koopman-factor-depth}
}
\caption{Comparison of same sample trajectories of the skeleton centroid estimated by (a) a vision-only approach, (b) the baseline factor graph system, and (c) the modified factor graph with the Koopman factor.
}
\label{fig:multiple-trajectories}
\end{figure}

However, the main contribution of the Koopman factor lies on the trajectory estimations at far distances. Figs.~\ref{fig:vision-only-depth} and \ref{fig:old-GTSAM-depth} show the evolution of the depth coordinate of the skeleton centroid for a few selected trajectories across multiple tests and participants. The vision-only trajectories start getting distorted around \unit[$3$]{m}, which is the range limit recommended by the manufacturer. While the baseline system was able to highly improve the skeleton estimated forms, it still struggles to track the walking kinematics of the skeleton. Fig.~\ref{fig:koopman-factor-depth} shows the same trajectories estimated by the new system with the Koopman factor. Its results highly contrast with the previous two approaches, with much smoother trajectories and no sudden jumps on the estimated depths. Paired with the results from Fig.~\ref{fig:all-boxplots}, they suggest that the new estimated skeleton trajectories preserve much better the nature of human walking, which largely amplifies the range of operation of the system to more than $10$ m.

\begin{figure}[h]
\centering
\subfloat[Without the Koopman factor]{\includegraphics[width=0.41\columnwidth]{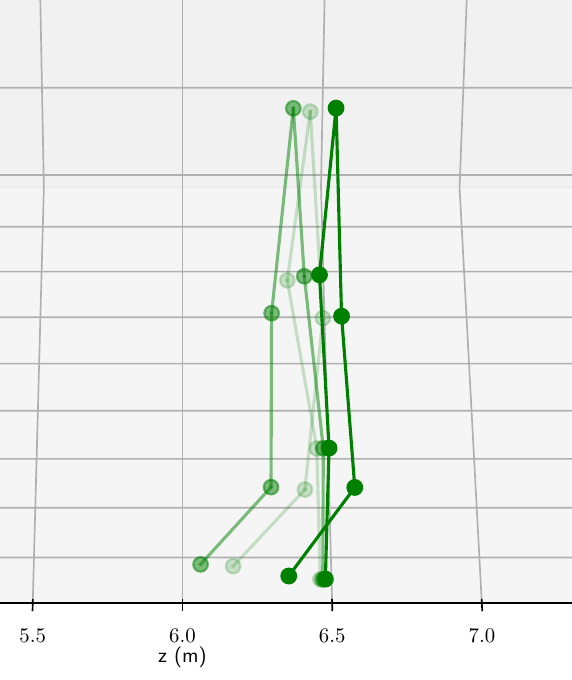}}
\hfill
\subfloat[With the Koopman factor]{\includegraphics[width=0.43\columnwidth]{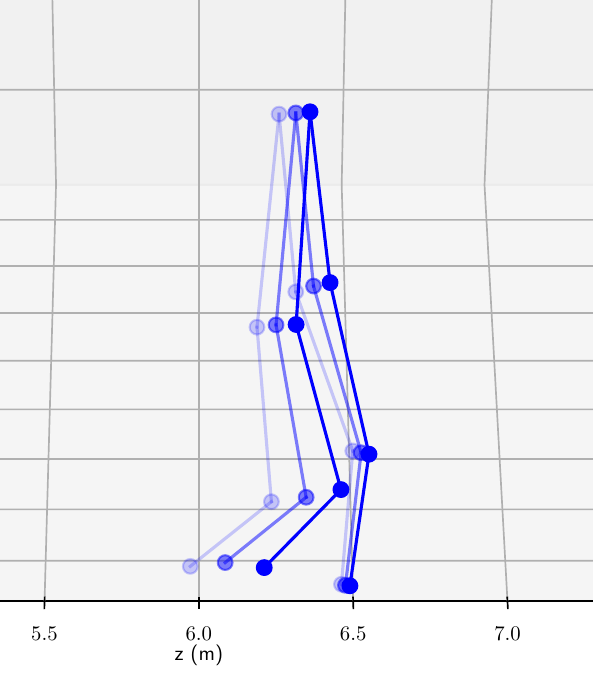}}
\caption{Three consecutive keyframes (with increasing opacity) corresponding to the side view of a skeleton walking. On the left, the estimations
obtained using the original factor graph system. On the right, the same keyframes estimated with the new Koopman factor.
}
\label{fig:skeleton-comparison}
\end{figure}

A more visual representation of this behavior can be seen in Fig.~\ref{fig:skeleton-comparison}, which compares the results of the system with and without the Koopman factor on three consecutive keyframes. While the depth joint coordinates of the original system's estimations fluctuate in an unnatural manner, the Koopman factor allows preserving the natural flow of the human walking away from the camera.

\section{CONCLUSIONS AND FUTURE WORK}
\label{sec:conclusions}


In this paper, we studied the viability of Koopman theory applied to the field of human pose recognition, targeted at the application of clinical assessment of lower limb mobility.

The novelty of our approach lies on the use of Koopman predictions of consecutive human skeleton keyframes within the optimization loop of a factor graph problem to influence the optimized skeleton estimations toward temporal consistency.
More specifically, we implemented a new factor graph factor that acts in parallel to the main IMU, image, depth, and contact factors of the original factor graph.
This new factor is applied on each individual keyframe of the trajectory, and runs a two-step process. First, a
trained spatial temporal graph convolutional network detects the walking activity the participant is performing among the selected set of \it{standing}, \it{sitting}, \it{standing up}, \it{sitting down}, and \it{walking}.
Then, this prediction is used to select a Koopman operator trained on the specified activity, and computes a measurement-free \it{a priori} skeleton estimation of the skeleton in the following keyframe.
This estimation is based on a data-driven learnt representation of the inherent dynamics of each one of the five selected walking activities,
and thus complements the information provided by the external measurements. Additionally, we validated the selection of the joint coordinates as features for the activity recognition system over an IMU-only alternative.

We studied the performance of the proposed Koopman factor by running our system against custom datasets composed of standard clinical tests for lower limb mobility assessment. Not only was the modified factor graph able to preserve the quality of the estimated skeletons observed in the original system, but it also showed a substantial improvement in filtering out the noise introduced by the depth sensor, extending the depth range of operation of the system to more than $10$ m away from the camera.

In future work, we plan to study more rigorous alternatives on the Koopman observables and frequencies selection, which should enable even better predictions without having to increase the Koopman dimensionality. Additionally, we plan to analyze the effect of the Koopman factor in those images where the human view is partially or completely blocked.
Finally, embedding the human walking dynamics into a Koopman representation could allow reducing the number of required IMUs. This fact would be of major importance toward building a clinical user-friendly system for automatic mobility assessment in the home.

\vspace{-1mm}



\bibliographystyle{IEEEtran}
\bibliography{biblio/icra2021.bib,biblio/websites.bib,biblio/iros2022mitjans.bib,biblio/ijrr2021VIFhumanWalking.bib}

\end{document}